\documentclass[letterpaper, 10 pt, journal, twoside]{IEEEtran}
\usepackage{amsmath,amsthm,amssymb}
\usepackage{graphicx}
\usepackage{xcolor}
\usepackage[ruled,vlined,linesnumbered]{algorithm2e}
\usepackage{centernot}
\usepackage{stackengine}
\usepackage{subfig}
\DeclareMathSizes{10}{10}{4}{4}
\usepackage{multirow}
\usepackage{makecell}

\usepackage{fancyhdr}
\fancypagestyle{plain}{
  \fancyhf{}
  \fancyhead{} 
  \fancyfoot{} 

}
\usepackage{eso-pic}
\usepackage{hyperref}

\theoremstyle{definition}

\newtheorem{theorem}{Theorem}
\newcommand{\indep}{\perp \!\!\! \perp}
\newcommand{\Lim}[1]{\raisebox{0.5ex}{\scalebox{0.8}{$\displaystyle \lim_{#1}\;$}}}

\ifCLASSINFOpdf
\else
\fi
\begin{document}
%

\AddToShipoutPictureBG*{
  \AtPageLowerLeft{
    \setlength\unitlength{1in}%
    \hspace*{\dimexpr0.5\paperwidth\relax}%
    \makebox(0,0.75)[c]{
      \parbox{0.7\paperwidth}{
        \raggedright\footnotesize
        © 2025 IEEE. Personal use of this material is permitted. 
        Permission from IEEE must be obtained for all other uses, in any current or future media, including reprinting/republishing this material for advertising or promotional purposes, creating new collective works, for resale or redistribution to servers or lists, or reuse of any copyrighted component of this work in other works.
      }
    }
  }
}

\title{An Alignment-Based Approach to Learning Motions from Demonstrations}
%
%
%

\author{Alex Cuellar$^{1}$, Christopher K. Fourie$^{1}$, and Julie A. Shah$^{1}$%
\thanks{Manuscript received: February, 11, 2025; Revised May, 21, 2025; Accepted August, 31, 2025.}
\thanks{This paper was recommended for publication by Editor Aleksandra Faust upon evaluation of the Associate Editor and Reviewers' comments.
This work was supported by the NSF grant on AI Coaching (IIS-2204914) and Los Alamos National Laboratories (grant C3464).
} 
\thanks{$^{1}$The authors are with the Massachusetts Institute of Technology, Cambridge, MA, 02139, USA
        {\tt\footnotesize alexcuel@mit.edu, ckfourie@csail.mit.edu, 	julie\_a\_shah@csail.mit.edu}}%
\thanks{Digital Object Identifier (DOI): \href{https://doi.org/10.1109/LRA.2025.3613956}{10.1109/LRA.2025.3613956}.}
}
%
%

\markboth{IEEE Robotics and Automation Letters. Preprint Version. SEPTEMBER, 2025}
{Cuellar \MakeLowercase{\textit{et al.}}: An Alignment-Based Approach to Learning Motions from Demonstrations} 

%



\maketitle

\begin{abstract}
Learning from Demonstration (LfD) has shown to provide robots with fundamental motion skills for a variety of domains. Various branches of LfD research (e.g., learned dynamical systems and movement primitives) can generally be classified into ``time-dependent” or ``time-independent” systems. Each provides fundamental benefits and drawbacks  -- time-independent methods cannot learn overlapping trajectories, while time-dependence can result in undesirable behavior under perturbation. This paper introduces Cluster Alignment for Learned Motions (CALM), an LfD framework dependent upon an alignment with a representative ``mean" trajectory of demonstrated motions rather than pure time- or state-dependence. We discuss the convergence properties of CALM, introduce an alignment technique able to handle the shifts in alignment possible under perturbation, and utilize demonstration clustering to generate multi-modal behavior. We show how CALM mitigates the drawbacks of time-dependent and time-independent techniques on 2D datasets and implement our system on a 7-DoF robot learning tasks in three domains.
\end{abstract}
﻿

%
\IEEEpeerreviewmaketitle
\section{Introduction} \label{sec:Intro}

\IEEEPARstart{A}{s} robots are introduced in industry and domestic settings, there is increasing need for robots to learn fundamental motions for given tasks. Learning from demonstrations (LfD) provides operators the ability to teach robots motions, ideally from a handful of demonstrations ($\sim$2 to 5)  \cite{ravichandar2020recent}. Such work allows robots to learn specific actions combining to complete more complex tasks. Existing methods can be partitioned into time-independent approaches, such as learned dynamical systems \cite{khansari2011learning, figueroa2018physically}, and time-dependent approaches, such as movement primitives \cite{ijspeert2013dynamical, paraschos2013probabilistic}. Despite advances in both areas, core limitations remain: time-independent methods cannot model overlapping trajectories, while time-dependent methods can exhibit undesirable behavior under perturbation -- if a robot is moved to another region of demonstrations, such methods have difficulty picking up where a perturbation leaves off \cite{ravichandar2020recent}. 

We introduce CALM (Cluster Alignment for Learned Motions), a novel LfD approach dependent on \textit{alignment} -- a mapping of states in the robot trajectory to those in one or more representative ``mean" trajectories for different clusters of demonstrations. CALM determines which mean trajectory to follow, and consistently updates its alignment to determine which section of the chosen mean trajectory to follow.  Additionally, if for any reason (e.g., a perturbation) the robot’s partial trajectory aligns better to another cluster’s mean trajectory, CALM will dynamically align to this new cluster and follow it (see Fig. 1). By consistently updating an alignment between the robot and demonstrated trajectories, CALM mitigates the limitations of time-dependent and -independent methods.

\begin{figure}
    \centering
    \includegraphics[width=\columnwidth]{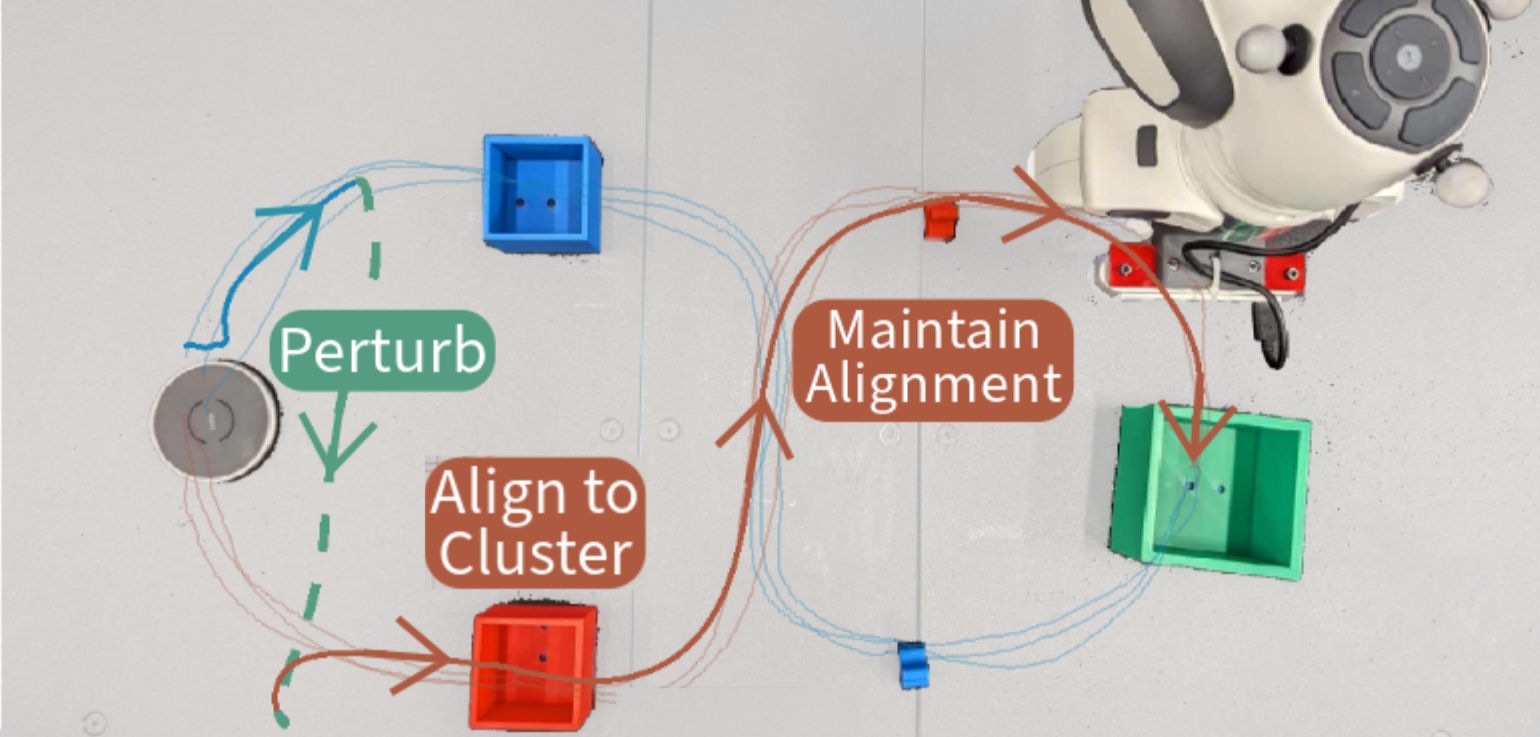}
    \caption{Example of CALM's learned behavior. There are two demonstration clusters. One cluster (\textit{blue}) passes the blue box and marker, while the other (\textit{red}) passes the red box and marker. The robot starts with \textit{blue}, but is perturbed and aligns to \textit{red}. CALM maintains alignment despite cluster overlap. }
    \label{fig:Franka_Teaser}
\end{figure}

While this paper introduces, to our knowledge, the first approach to LfD that uses partial trajectory alignment, techniques for trajectory alignment are not new \cite{dixon2005live, lasota2019bayesian}. However, existing methods generally assume (a) initial points of two trajectories align and (b) no sudden and large jumps in alignment. While reasonable in many settings, our need to maintain alignment during perturbations necessitates relaxing these constraints.

The primary contributions of this paper are as follows: 

\begin{enumerate}
    \item An alignment-dependent controller (CALM) for learning and reconstructing demonstrated motions.
    \item An HMM-based alignment method capable of handling alignment discontinuities from perturbation. 
    \item A method for choosing which of multiple demonstration clusters CALM follows given an alignment. 
\end{enumerate}

We also compare the CALM framework to time-dependent and time-independent approaches on various 2D datasets, and demonstrate CALM's viability in multiple robot domains. \footnote{Implementation and comparisons: https://github.com/AlexCuellar/CALM}

\section{Related Works} \label{sec:Related}

For ease of training, CALM is designed to learn motions from limited demonstrations ($\sim$2-5).  Therefore, we distinguish CALM from recent advances via large learning models such as diffusion \cite{chi2023diffusion} and action chunking transformers \cite{zhao2023learning}. Such methods show impressive capabilities in policy learning; however, even modern approaches require many demonstrations \cite{pmlr-v164-mandlekar22a}. Instead, we compare CALM to movement primitives, learned DS systems, and other lightweight methods effective when learned motions can be deployed more predictably. These methods can be partitioned into time-dependent and time-independent approaches, each subject to inherent limitations. 


\subsubsection{Time-Independent LfD}

Time-independent LfD takes the form of learned dynamical systems (DS) that map robot state $\mathbf{x}$ to desired velocity $\dot{\mathbf{x}}$. Many approaches are based in Khansari-Zadeh et al.'s SEDS framework, which guarantees  stability to a goal state \cite{khansari2011learning, figueroa2022locally}, while others learn a DS via neural networks \cite{lemme2014neural, urain2020imitationflow}. However, DS formulations fundamentally cannot represent trajectories that overlap with themselves. This is because DS's map each state $\mathbf{x}$ to one heading $\dot{\mathbf{x}}$.  In contrast, an overlapping trajectory must map the same state to different headings depending on which trajectory segment  is followed. 

\subsubsection{Time-Dependent LfD}
Time-dependent LfD methods largely center around Movement Primitives (MPs).  MPs are broadly classified into two categories, with extensions for various desired behaviors \cite{kulak2020fourier,li2023prodmp}. First, Dynamic Movement Primitives (DMPs) combine a learned forcing function with a linear attractor, providing stability grantees \cite{ijspeert2013dynamical}. However, DMPs only learn from one demonstration, making them inherently uni-modal. Probabilistic Movement Primitives (ProMPs) generate a distribution over trajectories which can be conditioned on a desired initial point or via-points \cite{paraschos2013probabilistic}. While probabilistic conditioning can produce multi-modal behavior, the lack of dynamics can lead to physically unrealistic trajectories and discontinuities \cite{li2023prodmp}. 

Others break from the MP structure entirely. 
Nawaz et al., for example, uses Neural ODEs to learn a DS (a time-independent formulation) augmented by a time-dependent correction that ensures stability and safety around obstacles \cite{nawaz2024learning}. Despite this time-dependent correction, the underlying DS means it has trouble with overlapping trajectories similar to purely time-independent formulations (see Section \ref{subsec:2DDatasets}). 

No matter the formulation, all time-dependent methods have potentially undesirable behavior when perturbed: while time-independent methods can adjust to a perturbation's endpoint, time-dependence causes trajectories to ``snap" back to the region dictated by the current timestep. To mitigate such behavior, others seek less strict time-dependency. Calilnon et al, for example, introduced a technique using Hidden Semi-Markov Models (HSMM) to determine which of several linear systems to follow \cite{calinon2011encoding,calinon2010learning}.  The HSMM depends both on state and time; thus the section of demonstrations followed by the system is informed by (but not strictly dependent on) time.  Therefore, if the robot is held in place for some time, it can continue smoothly once released even if time has passed.  However, this method is not provably stable, and large perturbations may cause unfamiliar transitions between attractors, confusing the learned transition function.  

Via alignment, CALM overcomes key limits: without enforcing one heading per state, it models overlapping motions, and time-independence enables natural perturbation recovery. By clustering demonstrations, CALM also captures multi-modal behavior that is stable about each mode’s endpoint.

\section{CALM Framework} \label{sec:Framework}

In contrast to prior work, Cluster Alignment for Learned Motions (CALM) determines a robot's velocity via an `alignment" between its trajectory and a mean of demonstrations. CALM is globally asymptotically stable in general, and globally asymptotically stable about demonstrations' endpoint (an important quality shared by prior methods \cite{ijspeert2013dynamical, figueroa2018physically}) given requirements on our alignment. We use Fourie et al.'s probabilistic TRACER technique to cluster demonstrations and generate a mean trajectory for each cluster \cite{fourie2024real}. TRACER is an Expectation Maximization algorithm that iteratively determines the probability of each trajectory corresponding to each mean trajectory, and updates mean trajectories to best represent the trajectories in its cluster.  Updates of the mean trajectory minimize the Dynamic Time Warping (DTW) cost between each mean and its corresponding trajectories.  TRACER can use a fixed number of clusters or infer one, letting operators decide whether to impose structure on demonstrations. 

We introduce the problem setting in Section \ref{subsec:Setting}. Then Section \ref{subsec:CALM_Framework} describes the control framework and Section \ref{subsec:stability} describes stability requirements. Finally, in Section \ref{subsec:velocity_matching}, we discuss how CALM produces motions which match the velocity profile of a mean trajectory. 

\subsection{Problem Setting} \label{subsec:Setting}
Let $\mathbf{x}^m$ be a mean trajectory of demonstrations and $\mathbf{x}^r$ be the robot's trajectory thus far. We model trajectories as sequences of states, $\mathbf{x} = [\mathbf{x}_1, \mathbf{x}_2, ... \mathbf{x}_n]$, each representing end effector coordinates $\mathbf{x}_i \in \mathbb{R}^d$. We notate the velocity at each state as $\dot{\mathbf{x}}_i \in \mathbb{R}^d$. States in $\mathbf{x}^m$ and $\mathbf{x}^r$ represent each trajectory at constant time intervals $\delta^m$ and $\delta^r$ respectively. 

Additionally, let alignment be represented by a ``task" variable, $\tau_k \in \mathbb{N}$, which maps state $\mathbf{x}^r_k$ in the robot’s trajectory to a state in the mean trajectory $\mathbf{x}^m$. This task variable indicates the robot trajectory's progress relative to the mean trajectory ($\tau_k = i$ indicates that state $\mathbf{x}^m_i$ best describes the robot's progress along the mean trajectory at state $\mathbf{x}^r_k$). Taking inspiration from prior work, we use a probabilistic interpretation of alignment \cite{lasota2019bayesian,fourie2024real}. $P(\tau_k = i | \mathbf{x}^r, \mathbf{x}^m)$ is the probability that $\mathbf{x}^r_k$ aligns to $\mathbf{x}^m_i$, with $p(\tau_k | \mathbf{x}^r, \mathbf{x}^m)$ being the corresponding distribution over all values of $\tau_k$.



\subsection{CALM Controller} \label{subsec:CALM_Framework}
In this paper, we formulate a controller as follows:
\begin{align}
    \label{eq:DS_Def}
    \Dot{\mathbf{x}} = k_v \frac{\nabla_{\mathbf{x}} g(\mathbf{x}, \mathbf{x}^r, \mathbf{x}^m)}{||\nabla_{\mathbf{x}} g(\mathbf{x}, \mathbf{x}^r, \mathbf{x}^m)||}
\end{align}
$\mathbf{x} \in \mathbb{R}^d$ is the robot state, and $k_v$ controls speed. Function $g$ uses the robot state and history ($\mathbf{x}^r$) to assess alignment with $\mathbf{x}^m$, and following $\nabla g$ moves the robot toward better alignment. Normalization in Eq.~\ref{eq:DS_Def} separates direction from speed, ensuring the velocity matches demonstrations. We define $g$ as: 
\begin{align}
    \label{eq:define_g}
    g(\mathbf{x}, \mathbf{x}^r, \mathbf{x}^m) = \sum_{i=1}^{F} q_i(\mathbf{x}) P(\tau_{k+1} = i | \mathbf{x}^r, \mathbf{x}^m)
\end{align}
Where $\mathbf{x}^r$ currently has $k$ states, $\mathbf{x}^m$ has $F$ states, and $q_{i}(\mathbf{x}) := p(\mathbf{x} | \tau_k = i, \mathbf{x}^m)$ is a Gaussian centered on $\mathbf{x}^r_i$ (thus, $q_i(\mathbf{x})$ increases when $\mathbf{x}$ is near $\mathbf{x}^m_i$):
\begin{align}
    \label{eq:define_obs}
    p(\mathbf{x} | \tau_k = i, \mathbf{x}^m)= \mathcal{N}(\mathbf{x} | \mathbf{x}^m_i, \Sigma_m )
\end{align}
Additionally, notice that $P(\tau_{k+1} = i | \mathbf{x}^r, \mathbf{x}^m)$ is an alignment probability. However, as $\mathbf{x}^r$ has $k$ states, $\tau_{k+1}$ represents the alignment between the robot's \textit{next} state and the mean trajectory. To calculate a distribution over the robot trajectory's next alignment, we introduce a static transition probability:
\begin{align}
    \theta_{j\rightarrow i} := P(\tau_{k+1}=i|\tau_k=j)
\end{align}
Using the assumption $\tau_{k+1} \indep \mathbf{x}^r, \mathbf{x}^m | \tau_k$: 
\begin{align}
    \label{eq:state_based_alignment}
    P(\tau_{k+1} \!=\! i | \mathbf{x}^r, \mathbf{x}^m) = \sum_{j=1}^{F} \theta_{j\rightarrow i} P(\tau_k = j | \mathbf{x}^r, \mathbf{x}^m)
\end{align}
Under the definition in Eq \ref{eq:define_g}, $g(\mathbf{x}, \mathbf{x}^r, \mathbf{x}^m)$ is a mixture of Gaussians, and its gradient is as follows: 
\begin{align}
    \label{eq:general_gradient_ln1}
    \nabla_{\mathbf{x}}g(\mathbf{x}, &\mathbf{x}^r, \mathbf{x}^m) = \sum_{i=1}^{F} \nabla_{\mathbf{x}} q_i(\mathbf{x}) P(\tau_{k+1} = i | \mathbf{x}^r, \mathbf{x}^m) \\
    \label{eq:general_gradient_ln2}
    &= \sum_{i = 1}^{F} \Sigma_m^{-1} (\mathbf{x}^m_i - \mathbf{x}) q_i(\mathbf{x}) 
    P(\tau_{k+1} = i | \mathbf{x}^r, \mathbf{x}^m)
\end{align}
Here, the gradient of $g$ will push the robot towards the points in $\mathbf{x}^m$ to which it will most likely align next. Consistently updating the robot alignment distribution $p(\tau_k|\mathbf{x}^r,\mathbf{x}^m)$ results in each gradient step pushing the robot further along demonstrations' mean trajectory $\mathbf{x}^m$. 

\subsection{Stability Guarantees} \label{subsec:stability}

Since $g$ is a mixture of Gaussians weighted by alignment $p(\tau_k|\mathbf{x}^r, \mathbf{x}^m)$, its gradient is globally asymptotically stable. Furthermore, requirements on the alignment probability can guarantee that the system is globally asymptotically stable about the final point of the mean trajectory $\mathbf{x}^m_{F}$: 
\begin{theorem} \label{thm:stability}
The CALM system defined in Eq \ref{eq:DS_Def} is globally asymptotically stable. \textit{(See proof in Appendix \ref{proof:thm1})}
\end{theorem}
\begin{theorem} \label{thm:alpha_convergence}
If $\Lim{k \rightarrow \infty} P(\tau_k = F | \mathbf{x}^r, \mathbf{x}^m) = 1$ and $\theta_{j\rightarrow i} = 0 \, \forall \, i < j$, the robot state will converge to the mean trajectory's final point:  $\mathbf{x} - \mathbf{x}^m_{F} \rightarrow 0$. \textit{(See proof in Appendix \ref{proof:thm2})}
\end{theorem}
For this paper, we use the transition function as follows: 
\begin{align} \label{eq:gradient_trans_fun}
    \theta_{j\rightarrow i} \propto
    \begin{cases}
        \phi(i,j + \Delta)  &  i \geq j\\
        0 & \text{otherwise } 
    \end{cases}
\end{align}
Where $\Delta = \delta^r/\delta^m$ is the ratio of time intervals between states in each trajectory and $\phi(\cdot, \cdot)$ is the radial basis function: 
\begin{align} \label{eq:rbf}
    \phi(\mathbf{a}, \mathbf{b}) = \exp{\left( -\frac{||\mathbf{a} - \mathbf{b}||^2}{2\sigma} \right)}
\end{align}
for a given $\sigma$. The transition probability is proportional to a radial basis function centered on a point slightly ahead of the current alignment (i.e., $\tau_k + \Delta$). However, each successive point of the robot trajectory cannot align to a prior point in the mean trajectory (i.e., $\theta_{j\rightarrow i}  = 0$ if $i < j$). This satisfies the transition function requirement in Theorem \ref{thm:alpha_convergence}.

\subsection{Velocity Matching} \label{subsec:velocity_matching}

To accurately replicate motions from demonstrations, the velocity gain $k_v$ in Eq \ref{eq:DS_Def} must match the speed of motions demonstrated. However, the robot should also ensure fast recovery from perturbations. Therefore, we formulate $k_v$ as a linear combination of an ``alignment" gain $k^a_v$ which matches the speed of demonstrations and a ``perturbed gain" $k^p_v$ which pulls the robot to the  mean trajectory if ever perturbed. We weigh $k^p_v$ heavily when the the robot position $\mathbf{x}$ is far from any mean trajectory position $\mathbf{x}_i^m$ and vice versa: 
\begin{align} \label{eq:vel_gain}
    k_v = \underset{i : ||\mathbf{x} - \mathbf{x}_{i}^m||}{\text{argmin}} \phi(\mathbf{x}, \mathbf{x}_{i}^m) k_v^a + (1 - \phi(\mathbf{x}, \mathbf{x}_{i}^m)) k_v^p
\end{align}
Where and $\phi(\cdot, \cdot)$ is the radial basis function. 

``Perturbed" gain $k^p_v$ is a hyperparameter, and we determine $k^a_v$ via the weighted sum of speeds from states in $\mathbf{x}^m$ based on alignment, as follows: 
\begin{align} \label{eq:align_vel_gain}
    k_v^a = \frac{\sum_{i=1}^{F} P(\tau_k = i | \mathbf{x}^r, \mathbf{x}^m) ||\dot{\mathbf{x}}_i^m||}{\sum_{i=1}^{F} P(\tau_k = i | \mathbf{x}^r, \mathbf{x}^m)}
\end{align}

\section{Alignment Technique for CALM} \label{sub:choices_of_alpha}

The CALM system described above requires an alignment probability distribution $p(\tau_k|\mathbf{x}^r,\mathbf{x}^m)$ updated for each newly observed state in the robot trajectory. This section describes an HMM alignment strategy capable of providing the desired perturbation and stability behavior for a CALM system. Section \ref{subsec:HMM} introduces the HMM alignment framework. Section \ref{subsec:transition_func} describes choices of HMM transition functions and how they impact system behavior. Finally, Section \ref{subsec:multiple_clusters} extends CALM's formulation to dynamically follow multiple trajectories while maintaining stability guarantees.

\subsection{Alignment Via HMM} \label{subsec:HMM}
We model alignment distributions $p(\tau_k|\mathbf{x}^r, \mathbf{x}^m)$ from Eq \ref{eq:state_based_alignment} as a hidden Markov model. We refer to each state in the robot trajectory as an ``observation," and alignment is our ``hidden" state. Our emission probability is the Gaussian $q_i(\mathbf{x})$ from Eq \ref{eq:define_obs}, and our transition probability is $\theta_{j\rightarrow i}$ (as we will discuss in Section \ref{subsec:transition_func}). We calculate the robot's alignment probability via the forward algorithm, where $\mathbf{x}^r_{1:k}$ is the robot's trajectory from state 1 to state $k$: 
\begin{multline}\label{eq:forward_alg}
    P(\tau_k = i,\mathbf{x}^r_{1:k}|\mathbf{x}^m) = q_i(\mathbf{x}^r_k) \\ \sum_{j=1}^{F} \theta_{j\rightarrow i}P(\tau_{k-1} = j, \mathbf{x}_{1:k-1}^r | \mathbf{x}^m)
\end{multline}
Where $P(\tau_{k-1}=j,\mathbf{x}^r_{1:k - 1}|\mathbf{x}^m)$ is remembered from the prior observation for every value $j \in \{1,...,F\}$. To retrieve $P(\tau_k = i | \mathbf{x}^r, \mathbf{x}^m)$ for Eq \ref{eq:state_based_alignment}, we calculate the following: 
\begin{align} \label{eq:conditionrobot}
    P(\tau_k=i|\mathbf{x}^r,\mathbf{x}^m) = \frac{P(\tau_k=i,\mathbf{x}^r|\mathbf{x}^m)}{\sum_{i=1}^F P(\tau_k=i,\mathbf{x}^r|\mathbf{x}^m)}
\end{align}
When calculating the HMM update in Eq \ref{eq:forward_alg}, we do not use Eq \ref{eq:gradient_trans_fun} for the transition function $\theta_{j\rightarrow i}$. Using the HMM approach, different choices of transition probability grant differing properties, such as convergence to the mean trajectory's final point and ability to be perturbed backwards along the mean trajectory.

\subsection{Choice of HMM Transition Function}
\label{subsec:transition_func}
Transition function $\theta_{j\rightarrow i}$ in Eq \ref{eq:forward_alg} defines how each new state in $\mathbf{x}^r$ alters its alignment to the mean $\mathbf{x}^m$. For the alignment method presented above to be provably asymptotically stable, we must follow the alignment condition in Theorem \ref{thm:alpha_convergence}. To satisfy this, we propose the following transition function: 
\begin{align} \label{eq:stable_trans_fun}
    \theta_{j\rightarrow i} \propto
    \begin{cases}
        \phi(i, j + \Delta)  &  i > j \\
        1 & i = j = F \\
        0 & \text{otherwise } 
    \end{cases}
\end{align}
This is similar to Eq \ref{eq:gradient_trans_fun}, with two differences. First, while Eq \ref{eq:gradient_trans_fun} requires $\theta_{j\rightarrow i}=0$ when $i < j$ (i.e., each new robot state cannot align to a prior point in the mean trajectory), Eq \ref{eq:stable_trans_fun} requires $\theta_{j\rightarrow i}=0$ when $i \leq j$ (i.e., that each new robot state must progress along the mean trajectory by at least one state). Second, Eq \ref{eq:stable_trans_fun} states that $\theta_{F \rightarrow F} = 1$, meaning once the robot aligns with the mean trajectory's final state, its alignment always remains at that final state. This transition function results in a CALM system stable about $\mathbf{x}^r_F$: 

\begin{theorem} \label{thm:HMM_stability}
    If a CALM system updates alignment probability $p(\tau_k|\mathbf{x}^r,\mathbf{x}^m)$ via the transition function in Eq \ref{eq:stable_trans_fun}, it converges to the final point in the mean trajectory: $\mathbf{x} - \mathbf{x}^m_{F} \rightarrow 0$. \textit{(See proof in Appendix \ref{proof:thm3})}
\end{theorem}

\textit{Alternate Transition Function:} The transition function in Eq \ref{eq:stable_trans_fun} can exhibit realignment when perturbations move the robot forward along the mean $\mathbf{x}^m$. However, it cannot realign to states already passed (i.e. it disallows $\theta_{j\rightarrow i} > 0$ when $i\leq j$). A transition function capable of such ``backwards" alignment is as follows, for a small value of $\epsilon$: 
\begin{align} \label{eq:back_trans_func}
    \theta_{j\rightarrow i} \propto
    \begin{cases}
        \phi(i, j + \Delta)  &  j \geq i\\
        \epsilon & \text{otherwise } 
    \end{cases}
\end{align}
Unlike Eq \ref{eq:stable_trans_fun}, this transition function allows $\theta_{j\rightarrow i} > 0$ when $i < j$. While global stability is not guaranteed, in Section \ref{subsec:BrushingTask} we empirically demonstrate consistent convergence while allowing ``backwards" perturbations. 
\subsection{Accounting for Multiple Demonstrated Clusters} \label{subsec:multiple_clusters}
The controller in Eq \ref{eq:DS_Def} recreates motions for one demonstrated task. However, TRACER can provide multiple mean trajectories if provided multiple demonstration clusters \cite{fourie2024real}. 
We use the HMM above to dynamically chose which mean trajectories to follow while maintaining stability guarantees. 

To achieve such behavior, we now assume $R$ TRACER mean trajectories $\{\mathbf{x}^{m_1} ,..., \mathbf{x}^{m_R}\}$. Additionally, we define a distribution describing how well the robot trajectory aligns to \textit{any} point in each mean trajectory via marginalizing the HMM joint probability in Eq \ref{eq:forward_alg}: 
\begin{align}
    \label{eq:marginalizetau}
    P(\mathbf{x}^r | \mathbf{x}^{m_i}) = \sum_{j=1}^{F_i} P(\tau_k = j, \mathbf{x}^r | \mathbf{x}^{m_i})
\end{align}
To account for multiple trajectories, we modify Eq \ref{eq:DS_Def} to follow the mean trajectory with the highest value of $p(\mathbf{x}^r | \mathbf{x}^{m_i})$: 
\begin{align}
    \label{eq:DS_Def_multi}
    \Dot{\mathbf{x}} = \underset{i^* : P(\mathbf{x}^r | \mathbf{x}^{m_{i}})}{\text{argmax}} k_v \frac{\nabla_{\mathbf{x}} g(\mathbf{x}, \mathbf{x}^r, \mathbf{x}^{m_{i^*}})}{||\nabla_{\mathbf{x}} g(\mathbf{x}, \mathbf{x}^r, \mathbf{x}^{m_{i^*}})||}
\end{align}

\begin{theorem}
    If a CALM system uses the transition function in Eq \ref{eq:stable_trans_fun}, as $k \rightarrow \infty$ it will converge to the endpoint of a cluster's mean trajectory: $\exists i \in \{1...R \} \; \mathbf{x}^r_k - \mathbf{x}^{m_i}_{F_{i}} \rightarrow 0$ \textit{(See proof in Appendix \ref{proof:thm4})}
\end{theorem}

\section{Experiments} \label{sec:Experiments}


In this section, we evaluate CALM on a suite of 2D environments in comparison to baselines with and without perturbations. Additionally, we evaluate CALM in three physical domains on a Franka robot in comparison to select baselines. 

We compare against five baselines: Probabilistic Movement Primitives (\textbf{ProMP}) \cite{paraschos2013probabilistic}, Dynamic Movement Primitives (\textbf{DMP}) \cite{ijspeert2013dynamical}, \textbf{CLF-NODE} \cite{nawaz2024learning}, \textbf{HSMM} \cite{calinon2011encoding}, and \textbf{LPV-DS} \cite{figueroa2018physically}. ProMP, DMP, and CLF-NODE generate time-dependent trajectories (ProMP generates a distribution over trajectories, and DMP and CLF-NODE learn time-dependent controllers) while LPV-DS learns a time-independent dynamical system. HSMM learns a transition function for attractors that is informed by (but not strictly dependent on) time. 

\subsection{Replication on 2D Datasets} \label{subsec:2DDatasets}

\begin{figure*}
    \centering
    \includegraphics[width=\textwidth]{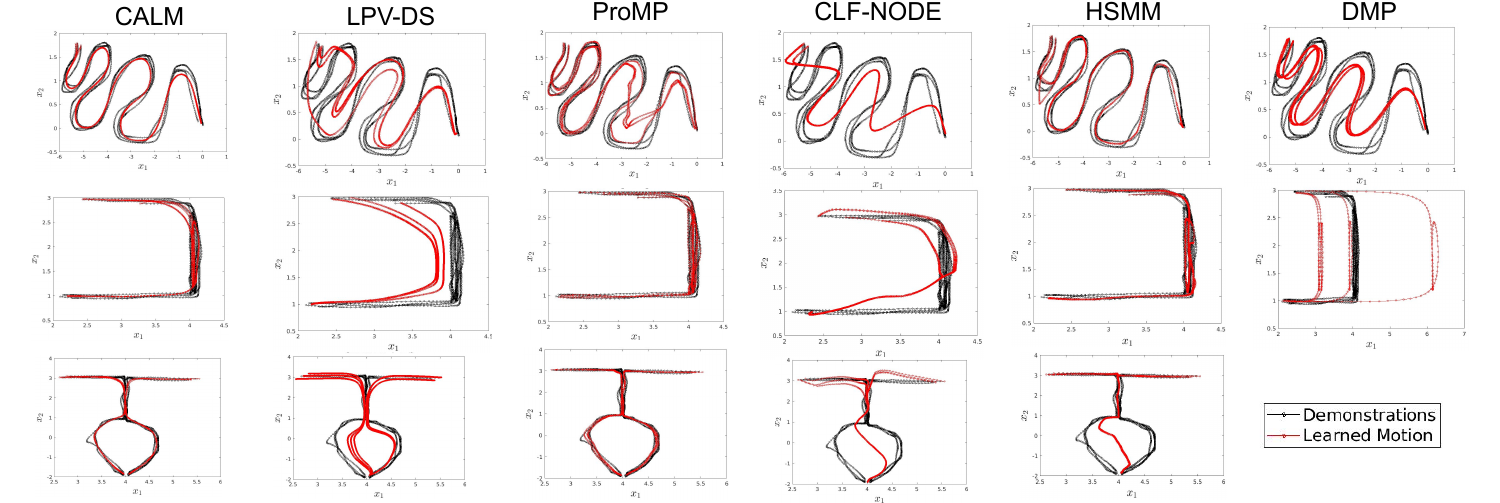}
    \caption{Behavior of each tested method on three datasets: \textit{Messy Snake}  (top), \textit{Overlap}  (middle), and \textit{Multi-Motion}  (bottom).}
    \label{fig:2D_Baselines}
\end{figure*}

\begin{figure}
    \centering
    \includegraphics[width=\columnwidth]{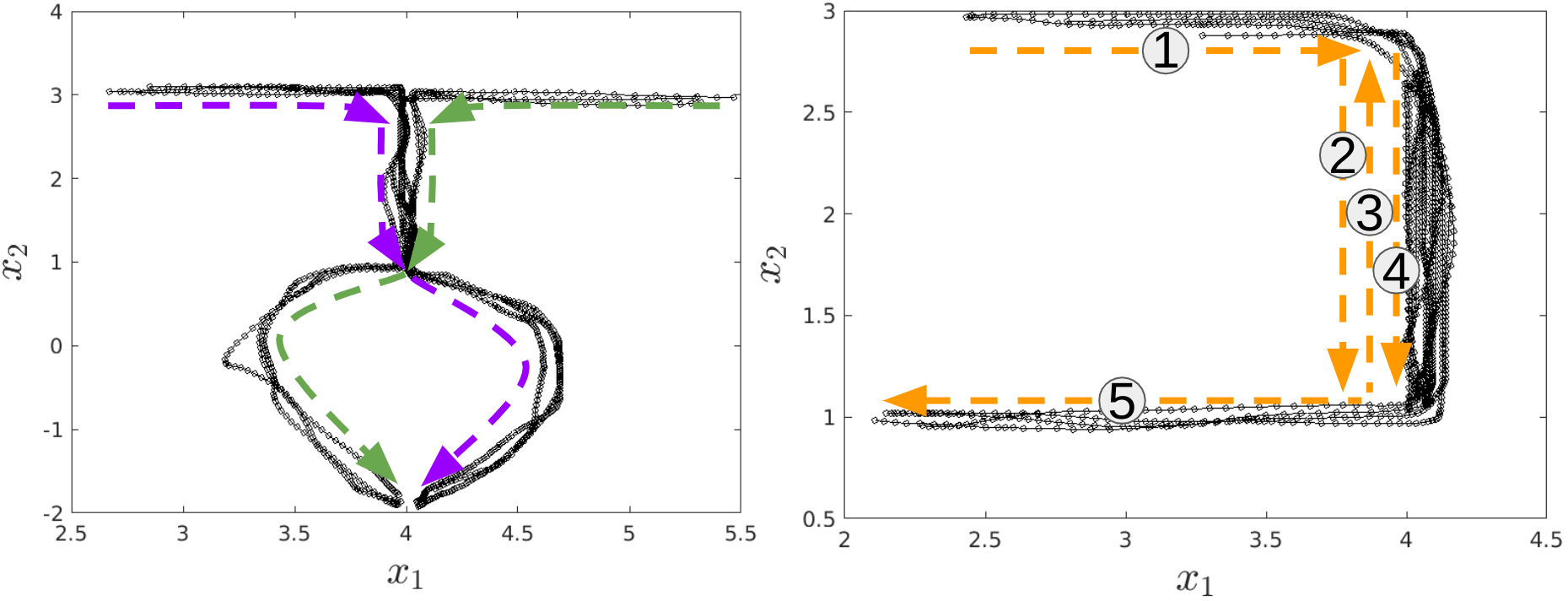}
    \caption{Two 2D motion datasets created for this paper: \textit{Multi-Motion} (left) and \textit{Overlap} (right). The Multi-Motion dataset includes six demonstrations following two clusters (shown in purple and green); Overlap includes four demonstrations which include a section of overlap (see numbered labels).}
    \label{fig:New_Datasets}
\end{figure}

We evaluate CALM via three illustrative 2D datasets. First, we use \textit{Messy Snake} from Figueroa et al. \cite{figueroa2018physically}. Second, we create \textit{Overlap}: four trajectories following a motion that overlaps itself. Finally, we create \textit{Multi-Motion}: six trajectories which follow one of two distinct but overlapping motions (Fig. \ref{fig:New_Datasets} depicts \textit{Overlap} and \textit{Multi-Motion}). 

Each method's performance is illustrated in Fig. \ref{fig:2D_Baselines}. For methods that learn a controller (CALM, DMP, LPV-DS, HSMM, and CLF-NODE) we begin at demonstrations' initial point and iteratively roll out a trajectory. For ProMP, we condition the trajectory's first point on demonstrations' initial point, and use the resulting distribution's mean as the predicted trajectory. We evaluate each method qualitatively and via the dynamic time warping distance (DTWD) between each demonstration and a generated motion initialized to the demonstration’s initial state. DTWD is a distance measure of trajectories’ spatial profiles used by prior work [3]. In MultiMotion, DTWD does not only evaluate methods’ ability to follow demonstrations generally, but whether they consistently follow the correct cluster given initial states and perturbations.

We first qualitatively evaluate each method (see Fig. \ref{fig:2D_Baselines}): 

\textit{CALM:} CALM's alignment allows it to follow the overlaps in \textit{overlap}, and representing clusters allows it follow each motion in \textit{multi-motion}. TRACER's ability to model intricate demonstrations allows it to follow the twists of \textit{messy-snake}. 

\textit{LPV-DS:} This method's time-independence means that it cannot model the overlaps in \textit{overlap}. Since it does not model trajectory clusters, it starts following the wrong cluster when the two motions in \textit{multi-motion} overlap. 

\textit{ProMP:} ProMP's basis functions struggle to model \textit{messy-snake} (more basis functions help, but performance plateaued at 100).  Its time-dependence allows it to model \textit{overlap}, and conditioning on initial state allows it to follow each motion in \textit{multi-motion} (despite no explicit cluster representation).

\textit{CLF-NODE:} Despite a time-dependent correction, CLF-NODE's underlying time-independent DS cannot model the overlaps in \textit{overlap}. Its underlying neural DS also struggles to capture the twists of \textit{messy-snake}, and its lack of modeling trajectory clusters results in a compromise between the two motions in \textit{multi-motion}, cutting between the two. 

\textit{HSMM:} HSMM's time-informed (but not entirely time-dependent) transition function between attractors means that it can model \textit{overlap}, and it can model the twists in \textit{messy-snake} (except a bit around the first turn).  However, similar to CLF-NODE, it cuts between the two clusters of \textit{multi-motion} because it does not model demonstration clusters.

\textit{DMP:} Similar to ProMP, DMPs' basis functions struggle to model the complexity of \textit{messy-snake}.  DMPs' forcing function is also sensitive to similar start and end states along one dimension, causing warping along $x_1$ in \textit{overlap}. DMPs can only learn from one trajectory, and cannot learn \textit{multi-motion} at all (we use TRACER's mean for other datasets). 

\begin{table}[t]
        \centering
            \begin{tabular}{c|c|c|c|c|c|c} 
                 & CALM & LPV-DS & ProMP & NODE & HSMM & DMP\\\hline
                \textit{MS} & 48.48 & 196.48 & 76.34 & 186.59 & 46.95 & 93.19\\
                \textit{O} & 17.12	& 139.77 & 8.51 & 92.11 & 55.98  & 310.5\\
                \textit{MM} & 12.77 & 87.64 & 8.42 & 70.24  & 15.60 & NA\\
            \end{tabular}
            \caption{Mean DTWD across methods (Dataset labels: \textit{MS} = \textit{Messy Snake}, \textit{O} = \textit{Overlap}, \textit{MM} = \textit{Multi-Motion}). CALM, ProMP, and HSMM generally outperform other methods. }\label{table:dtw_errors}
        \end{table}
        
Table \ref{table:dtw_errors} shows the mean DTWD for each method on each dataset. ProMP, HSMM, and CALM outperform the other methods across all datasets, but CALM outperforms HSMM on \textit{Multi-Motion} and ProMP on \textit{Messy Snake}. On \textit{Overlap} and \textit{Multi-Motion}, CALM's tendency to approach a mean trajectory increased errors compared with ProMP, which is generally able to follow each demonstration's specific trajectory. However, Fig. \ref{fig:2D_Baselines} shows little difference between the generated trajectories in with \textit{Overlap} and \textit{Multi-Motion}. 

\subsection{Behavior Under Perturbation} \label{subsec:Perturbation}

\begin{figure*}
    \centering
    \includegraphics[width=.97\textwidth]{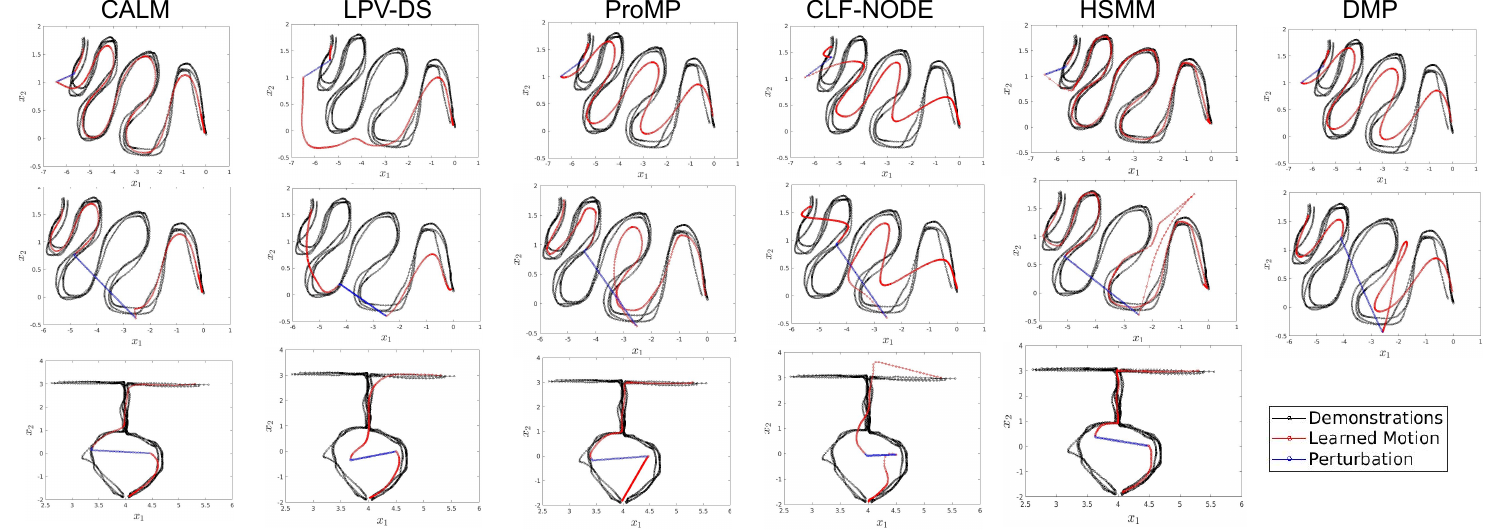}
    \caption{Behavior under perturbation: \textit{Messy Snake} with ``uninformative" perturbation (top), \textit{Messy Snake} with ``informative" perturbation to another region of demonstrations (middle), \textit{Multi-Motion} with ``informative" perturbation (bottom). }
    \label{fig:2D_Baselines_Perturb}
\end{figure*}

This section compares CALM's behavior to the baselines' when given perturbations. Note that ProMP learns a distribution over trajectories rather than a controller that is iteratively rolled out; thus, ProMP has no defined way to handle a perturbation. For these experiments, we assumed a controller following a trajectory generated from a ProMP would always move toward $\mathbf{x}_i$ at timestep $i$ while matching the generated trajectory's speed at timestep $i$. Fig. \ref{fig:2D_Baselines_Perturb} highlights the perturbation behavior for CALM and each baseline. We distinguish ``uninformative" perturbations (those not moving the state into a separate region of demonstrations) from ``informative" perturbations (those that move the state into a separate region of demonstrations). As the distinction lies in user intent, it is subjective by design -- ``uninformative" perturbations may involve accidental bumps, while ``informative" perturbations represent users intentionally skipping or repeating sections of a trajectory, or changing which trajectory cluster to follow. 

All methods except LPV-DS recover from ``uninformative" perturbations, as shown in the top row of Fig. \ref{fig:2D_Baselines_Perturb}. This is because the DS tends to model the contours of a trajectory less well when further from demonstrations. 
The time-dependence inherent to DMP, ProMP, HSMM, and CLF-NODE make them move backward along the demonstrations when given an ``informative" perturbation, while the time-independence of CALM and LPV-DS are able to pick up where the perturbation left off. Similarly, for ``informative" perturbations on \textit{Multi-Motion} that shift from one demonstration cluster to another, CALM can realign with the new demonstration cluster because it models each cluster explicitly, while ProMP and CLF-NODE do not model each cluster and cannot realign as CALM does. 

\begin{figure}
    \centering
    \includegraphics[width=.95\columnwidth]{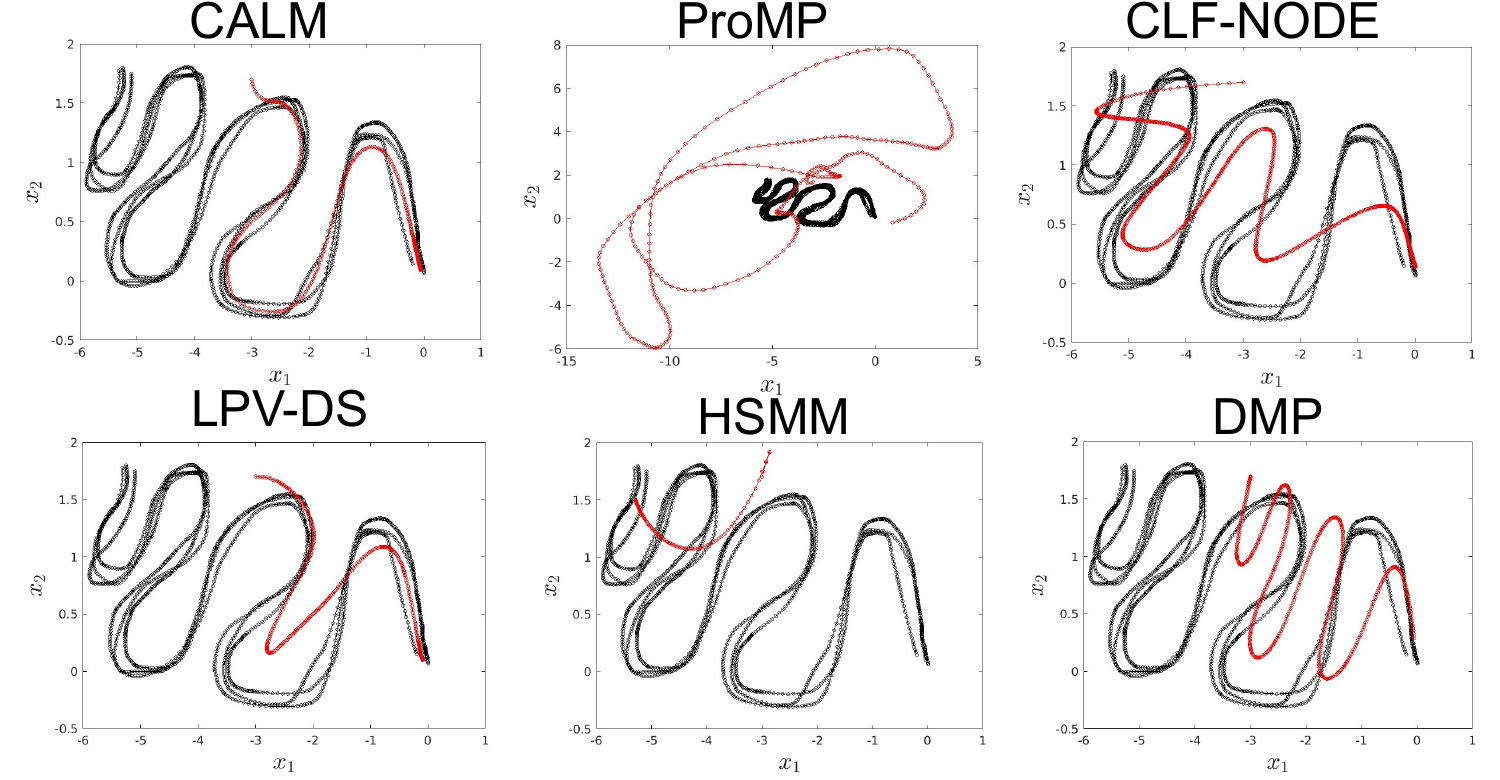}
    \caption{Each method with an out-of-distribution initial state.}
    \label{fig:2D_Baselines_Perturb_Diff_Init}
\end{figure}

Fig. \ref{fig:2D_Baselines_Perturb_Diff_Init} depicts methods given unfamiliar initial states. CALM and LPV-DS are time-independent and can begin their motion in the middle of demonstrations near the initial state, while time-dependent methods cannot. DMP's forcing function performs the full motion warped to the initial state. CLF-NODE and HSMM return to the demonstrations' beginning based on the current timestep, and ProMP does not achieve any coherent imitation of demonstrations. ProMPs can handle conditioning on initial positions near those demonstrated \cite{paraschos2013probabilistic}. However, out-of-distribution initial states force unfamiliar trajectories which ProMP conditioning cannot model, causing erratic behavior.  

\subsection{Robot Experiments} \label{subsec:BrushingTask}

We apply CALM to three kinesthetically taught tasks on a 7-DoF Franka robot: wiping, writing, and repetitive brushing. The robot uses a compliant PI torque controller using the pseudo-inverse Jacobian to translate desired velocity to torque. We also implement one time-dependent method (ProMP, which had the lowest DTWD in Table \ref{table:dtw_errors}) and one time-independent method (LPV-DS). This paper's associated video depicts the robot’s behavior in all tasks. \footnote{See video at: https://youtu.be/qOFSgZ0K-eo}

\begin{figure}
    \centering
    \includegraphics[width=\columnwidth]{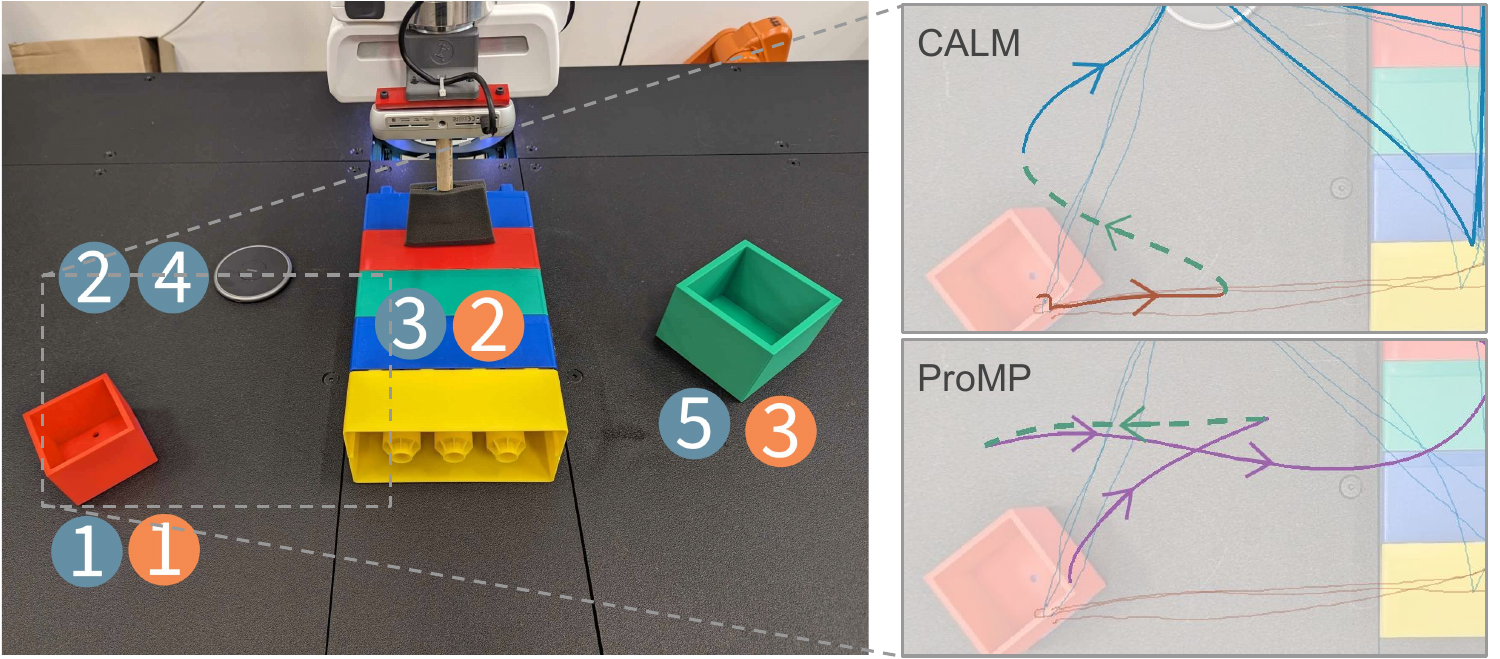}
    \caption{Left shows our setup and each variant's numbered steps. The ``standard" variant (red) starts at the red box, wipes the object, and ends at the green box; ``inspection" (blue) stops at the inspection station before and after wipes. Top right shows CALM on a portion of the motion: the robot begins following ``standard" (red), is perturbed (green), then realigns to ``inspection" (blue). Bottom right shows ProMP: it is unable to distinguish between the variants before or after perturbation, and instead cuts between the two (purple). }
\label{fig:brushing}
\end{figure}

\subsubsection{Wiping Task} We provide six demonstrations of a wiping task representative of motions to detect contaminants on an object. We demonstrate two task variants (Fig. \ref{fig:brushing}). In the first, ``standard" variant, the robot picks up a brush, wipes the object twice, and moves it to a disposal point. The second variant, ``inspection," requires the robot to move the brush to a detector before and after brushing. The variants overlap significantly with themselves and each other, and realignment from intentional perturbation between variants must be maintained for proper task execution. For this task and the writing domain, the robot uses the ``backwards" transition function from Section \ref{subsec:transition_func}. Despite no provable convergence to demonstration endpoints, both tasks empirically converge. CALM is able to maintain alignment, and switch between variants when desired.  Neither ProMP or LPV-DS represent demonstration clusters, and cannot distinguish the two variants: ProMP combines elements of each, while LPV-DS fails to model ``inspection" entirely. LPV-DS's time-independence results in failure to model overlaps in the trajectory, and only wipes the object once (see video). 

\subsubsection{Robot Writing} 
\begin{figure}
    \centering
    \includegraphics[width=.97\columnwidth]{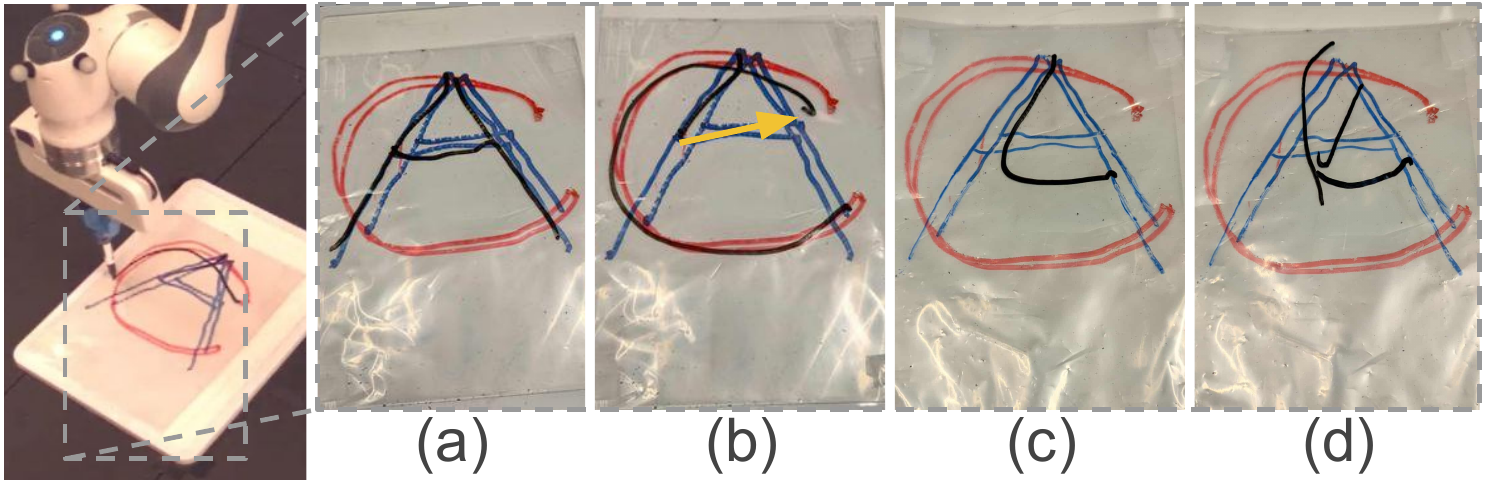}
    \caption{Demonstrations writing ``C" (red), ``A" (blue), and the robot motions (black). (a) CALM drawing ``A" (b) CALM with perturbation from ``A" to ``C" (c) LpvDS (d) ProMP}
\label{fig:Robot_Writing}
\end{figure}

We teach the robot to write two letters: A and C. Robot writing is a common baseline in LfD \cite{figueroa2022locally}. However, we demonstrate in the same region of the state space, showing CALM's capability to maintain alignment in overlapping multi-modal tasks. Fig. \ref{fig:Robot_Writing} shows CALM's reconstruction of each letter and ability to be perturbed from one letter to another.  LPV-DS and ProMP's inability to model demonstration clusters renders both unable to write either letter, let alone realign after informative perturbation. 
\subsubsection{Repetitive Brushing} 
\begin{figure}
    \centering
    \includegraphics[width=\columnwidth]{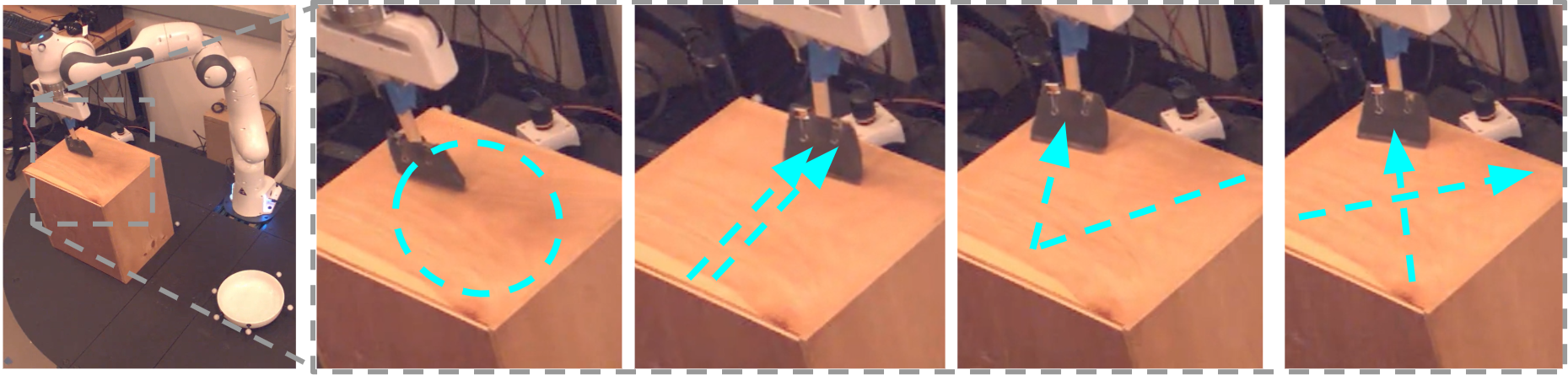}
    \caption{Four brush strokes learned by CALM. An altered transition function allows for indefinite repetitions of each.}
\label{fig:Repetative_Brushing}
\end{figure}

Here, we teach the Franka to dip a brush into a ``paint" bowl and brush a box.  We teach four strokes that are repeated without a defined endpoint (Fig. \ref{fig:Repetative_Brushing}). Learning repeated motions is a common extension of other LfD methods \cite{petrivc2014online,nawaz2024learning}, and we use this to demonstrate how modified transition functions can produce desired behavior beyond those described in Section \ref{subsec:transition_func}. For this repeated behavior, we use the modified transition function: 
\begin{align} \label{eq:periodic_trans_func}
    \theta_{j \rightarrow i} \propto
    \begin{cases}
        \phi(i, j + \Delta)  &  i > j\\
        1 & j = F \text{\, and \,} i = 1 \\
        \epsilon & \text{otherwise } 
    \end{cases}
\end{align}

CALM consistently maintains alignment to a single stroke, and reliably switches between different strokes with intentional perturbations despite four clusters of demonstrations (as opposed to two in prior tasks). Such behavior demonstrates the flexibility offered by modified transition functions while maintaining desired properties. We do not train LPV-DS or ProMP on this domain, as neither can learn repeated behaviors. See the associated video for example behavior.

\section{Limitations and Future Work} \label{sec:FutureWork}

This paper presents a framework for alignment-based LfD; however, several capabilities are left to future work. First, many methods allow for trajectory modulation to satisfy via-points or modified end goals \cite{ravichandar2020recent, paraschos2013probabilistic}. Future work may take inspiration from Berio et al.'s use of splines in LfD \cite{berio2017generating}: approximating and manipulating splines on mean trajectories may offer the ability to achieve desired via-point modulation. 

Second, we focused on motions in Cartesian space. Prior work has shown the benefits of other coordinate systems in domains such as manipulation \cite{ti2023geometric}. Similar extensions may be a promising direction, but outside this paper's scope. 

Lastly, LfD conditioned on sensory data \cite{li2023prodmp} and language input \cite{sharan2024plan} has shown capabilities in more complex behavior. Similar conditioning could inform CALM's choice of mean trajectory and where to align within a mean trajectory. 

\section{Conclusion} \label{sec:Conclusion}

This paper presented CALM, a learning from demonstrations framework based on alignment. First, we introduced the gradient-based methodology and convergence properties. We then introduced a partial-trajectory alignment technique via HMMs to allow large jumps and convergence to multiple means. Finally, we demonstrated CALM's improvement over baseline methods with regard to tracking, overlapping behavior, and realignment after perturbation on 2D datasets and multiple real robot domains. 



\appendices

\section{proofs}
\subsection{Theorem 1} \label{proof:thm1}
Consider Eq \ref{eq:general_gradient_ln2}, rewritten in Eq \ref{eq:ds_new} with $c_i(\mathbf{x},\tau_{k+1}=i) := q_i(\mathbf{x}) 
P(\tau_{k+1} = i | \mathbf{x}^r, \mathbf{x}^m)$. Note that $A_i \prec 0 \, \forall i \in \{1,\ldots, F\} \implies \sum A_i = A \prec 0$ 
(i.e. the sum of negative definite matrices is negative definite) and $\sum c_i > 0$. Hence, the system is always globally asymptotically stable (G.A.S.) around an attractor defined by $\tau_{k+1}$. :
\begin{align}
    \label{eq:ds_new}
    \nabla_\mathbf{x} g(\mathbf{x},\mathbf{x}^r,\mathbf{x}^m) &= \sum_{i = 1}^{F} \underbrace{\Sigma_m^{-1}}_{\succ 0} (\mathbf{x}^m_i - \mathbf{x})\underbrace{c_i(\mathbf{x},\tau_{k+1}=i) }_{\geq 0}\\
    &= \sum_{i = 1}^{F} \left( A_i \mathbf{x} + \mathbf{b}_i  \right) = A \mathbf{x} + \mathbf{b} 
\end{align}
Note that the global attractor $\mathbf{x}^*$ is:
\begin{align}
    \label{eq:global_attractor}
    \mathbf{x}^* = -(A^{-1}\mathbf{b}) = \frac{\sum_i c_i \mathbf{x}^m_i}{\sum_i c_i}
\end{align}
Thus, if $P(\tau_{k+1} = i|\mathbf{x}^r,\mathbf{x}^m) = 1$, the attractor $\mathbf{x}^* = \mathbf{x}^m_i$.
\subsection{Theorem 2} \label{proof:thm2}
If our alignment strategy guarantees $P(\tau_k = F | \mathbf{x}^r, \mathbf{x}^m) = 1$, then as $k \rightarrow \infty$, Eq \ref{eq:state_based_alignment} reduces to the following: 
\begin{align} \label{eq:state_based_converge_reduced}
    P(\tau_{k+1} = i | \mathbf{x}^r, \mathbf{x}^m) = \theta_{F\rightarrow i}P(\tau_k = F | \mathbf{x}^r, \mathbf{x}^m)
\end{align}
If $\theta_{j\rightarrow i} = 0 \; \forall i < j$, notice $P(\tau_{k+1} = F | \mathbf{x}^r, \mathbf{x}^m) = 1$. Thus, by Eq \ref{eq:global_attractor}, the DS in Eq \ref{eq:DS_Def} is stable about $\mathbf{x}^m_{F}$. 

\subsection{Theorem 3} \label{proof:thm3}
We prove this theorem by demonstrating that the condition $P(\tau_{k}=F|\mathbf{x}^r,\mathbf{x}^m) = 1$ from Theorem \ref{thm:alpha_convergence} is satisfied when using the transition function in Eq \ref{eq:stable_trans_fun}. We show via induction that a robot trajectory $\mathbf{x}^r$ of length $k$ necessitates that $P(\tau_{k} = i, \mathbf{x}^r_{1:k} | \mathbf{x}^m) = 0 \; \forall i < k$: 

\textbf{Base case}: Assume $P(\tau_{k-1}=i, \mathbf{x}^r_{1:k-1} | \mathbf{x}^m)$ is non-zero for all values $i = \{1,...,F\}$. Using the transition function from Eq \ref{eq:stable_trans_fun}, we have $\theta_{i\rightarrow 1} = 0$ for any value $i$. Therefore, by Eq \ref{eq:forward_alg}, $P(\tau_k = 1,\mathbf{x}^r_{1:k}|\mathbf{x}^m) = 0$. 

\textbf{Inductive Case}: Assume $P(\tau_{k-1} = j, \mathbf{x}_{1:k-1}^r | \mathbf{x}^m) = 0 \; \forall j < v$ for some $v \geq 1$. In the next HMM update, we can remove the first $v$ elements of the sum in Eq \ref{eq:forward_alg}, leaving: 
\begin{multline}
P(\tau_k=i,\mathbf{x}^r_{1:k}|\mathbf{x}^m) = q_i(\mathbf{x}_k^r) \\ \sum_{j=v}^{F} \theta_{j\rightarrow i} P(\tau_{k-1} = j, \mathbf{x}_{1:k-1}^r | \mathbf{x}^m)
\end{multline}
For any $j \in \{v,...,F\}$, $\theta_{j\rightarrow i} = 0$ when $i < v + 1$. Thus, $P(\tau_k=i,\mathbf{x}^r_{1:k}|\mathbf{x}^m) = 0 \; \forall i < v+1$, completing the induction. 

Once $v = F$, we have $P(\tau_{k-1}=j,\mathbf{x}^r|\mathbf{x}^m) = 0 \; \forall j < F$, leaving the HMM update as follows: 
\begin{multline} \label{eq:HMM_to_infity}
    P(\tau_k \!= \! i,\mathbf{x}^r|\mathbf{x}^m) \!= \! q_i(\mathbf{x}^r_k) \theta_{F\rightarrow i} P(\tau_{k-1} \!= \! F, \mathbf{x}_{1:k-1}^r | \mathbf{x}^m)
\end{multline}
Since $\theta_{F\rightarrow F} = 1$, $P(\tau_k = i,\mathbf{x}^r| \mathbf{x}^m) = 0 \; \forall i\neq F$. Via Eq \ref{eq:conditionrobot}, $P(\tau_k = F|\mathbf{x}^r, \mathbf{x}^m)=1$, satisfying the condition in Thm \ref{thm:alpha_convergence}. 

\subsection{Theorem 4} \label{proof:thm4}
\noindent \textbf{Lemma 1:} As $k \! \! \rightarrow \! \! \infty$, $P(\mathbf{x}^r_{1:k+1}|\mathbf{x}^{m_i}) \! = \! q_{F_i}\!(\mathbf{x}^r_{k+1}) P(\mathbf{x}^r_{1:k}|\mathbf{x}^{m_i})$ \\
\textit{Proof: } Eq \ref{eq:HMM_to_infity} in Thm \ref{thm:HMM_stability} shows that as $k\rightarrow \infty$: 
\begin{align}
    &\begin{aligned}
    P(\tau_{k+1} \! \! = \! j,\mathbf{x}^r|\mathbf{x}^{m_i} \!) \! \! = \! \! q_j(\mathbf{x}^r_{k+1}) \theta_{F_i\rightarrow j} 
    P( \! \tau_k \! = \! \! F_i, \! \mathbf{x}_{1:k}^r | \mathbf{x}^{m_i} \! ) 
    \end{aligned}
    \\
    & \; =
    \begin{cases}
        q_j(\mathbf{x}^r_k)P(\tau_k = j, \mathbf{x}_{1:k}^r | \mathbf{x}^m) & j = F_i \\
        0 & \text{Otherwise}
    \end{cases}
\end{align}
Where step 2 uses Eq \ref{eq:stable_trans_fun}. Substituting into Eq \ref{eq:marginalizetau} and observing via Theorem \ref{thm:HMM_stability} that $P(\mathbf{x}^r_{1:k}|\mathbf{x}^{m_i}) = P(\tau_k = F_i, \mathbf{x}^r_{1:k} | \mathbf{x}^{m_i})$: 
\begin{align}
    P(\mathbf{x}^r_{1:k+1} | \mathbf{x}^{m_i}) &= q_{F_i}(\mathbf{x}^r_k)P(\tau_k = F_i, \mathbf{x}^r_{1:k} | \mathbf{x}^{m_i})\\
    &= q_{F_i}(\mathbf{x}^r_k)P(\mathbf{x}^r_{1:k}|\mathbf{x}^{m_i}) \; \hspace*{0pt}\hfill \qedsymbol{}
\end{align}
\noindent \textbf{Lemma 2} As $k\rightarrow \infty$, $\nabla_\mathbf{x}g(\mathbf{x},\mathbf{x}^r,\mathbf{x}^{m_i}) \propto \nabla_\mathbf{x}q_{F_i}(\mathbf{x})$ \\
\textit{Proof:} Using Theorem \ref{thm:alpha_convergence}, as $k \rightarrow \infty$, Eq \ref{eq:define_g} becomes:
\begin{align}
    g(\mathbf{x},\mathbf{x}^r,\mathbf{x}^{m_i}) = q_{F_i}(\mathbf{x})p(\tau_{k+1}=F_i|\mathbf{x}^r,\mathbf{x}^{m_i}) \;\hspace*{0pt}\hfill \qedsymbol{}
\end{align}
Let $i^*$ be the index of the mean trajectory for which: 
\begin{align} \label{eq:thm4idominance}
p(\mathbf{x}^r|\mathbf{x}^{m_{i^*}}) > p(\mathbf{x}^r|\mathbf{x}^{m_i}) \; \forall i \neq i^*
\end{align}
At time $k-1$, let $i^*=i^\prime$. At time $k$,
$i^*$ can either remain the same ($i^* = i^\prime$) or change ($i^* \neq i^\prime$). 
If $i^*$ remains the same, Eq \ref{eq:DS_Def} and Lemma 2 indicate $q_{F_{i^\prime}}(\mathbf{x}^r_{k}) \geq q_{F_{i^\prime}}(\mathbf{x}^r_{k-1})$.
Equality only occurs at the max of $q_{F_{i^\prime}}(\cdot)$ where $\nabla q_{F_{i^\prime}}(\mathbf{x}^r_k) = 0$: 
\begin{align} \label{eq:thm4stablept}
    \mathbf{x}^r_k = \mathbf{x}^{m_{i^\prime}}_{F_{i^\prime}}
\end{align}
Now imagine at time $k$, $i^*\!=\!i^{\prime\prime}$ ($i^*$ changes). Here, $\mathbf{x}^{m_{i^{\prime\prime}}}$ ``overtakes" $\mathbf{x}^{m_{i^\prime}}$ so that $p(\mathbf{x}^r|\mathbf{x}^{m_{i^{\prime\prime}}}) \!>\! p(\mathbf{x}^r|\mathbf{x}^{m_{i^\prime}})$. Notice by Lemma 1, this ``overtaking" requires $q_{F_{i^{\prime\prime}}}(\mathbf{x}^r_k) \!>\! q_{F_{i^\prime}}(\mathbf{x}^r_k)$. Therefore, whether $i^*$ changes or not, $q_{F_{i^*}}(\mathbf{x}^r_k)$ increases at every step.  Thus, as $k$ increases, $\mathbf{x}_k^r$ must eventually reach a max of $q_{F_{i^*}}(\cdot)$ where $\nabla q_{F_{i^*}}(\mathbf{x}^r_k) = 0$, as in Eq \ref{eq:thm4stablept}. By Lemma 2, $\nabla q_{F_{i^*}}(\mathbf{x}^r_k) = 0 \implies \nabla g(\mathbf{x}^r_k,\mathbf{x}^r,\mathbf{x}^{m_i}) = 0$. Thus, $\nabla g$ is stable at point $\mathbf{x}^{m_i}_{F_i}$ for some $i$, proving the theorem.


\bibliographystyle{IEEEtran}
\bibliography{main.bib}

@preamble{ "\newcommand{\noopsort}[1]{} "
	# "\newcommand{\printfirst}[2]{#1} "
	# "\newcommand{\singleletter}[1]{#1} "
	# "\newcommand{\switchargs}[2]{#2#1} " }

@article{figueroa2018physically,
  title={A physically-consistent bayesian non-parametric mixture model for dynamical system learning},
  author={Figueroa Fernandez, Nadia Barbara and Billard, Aude},
  journal={Proceedings of Machine Learning Research},
  year={2018}
}

@article{paraschos2013probabilistic,
  title={Probabilistic movement primitives},
  author={Paraschos, Alexandros and Daniel, Christian and Peters, Jan R and Neumann, Gerhard},
  journal={Advances in neural information processing systems},
  volume={26},
  year={2013}
}

@article{ijspeert2013dynamical,
  title={Dynamical movement primitives: learning attractor models for motor behaviors},
  author={Ijspeert, Auke Jan and Nakanishi, Jun and Hoffmann, Heiko and Pastor, Peter and Schaal, Stefan},
  journal={Neural computation},
  volume={25},
  number={2},
  pages={328--373},
  year={2013},
  publisher={MIT Press One Rogers Street, Cambridge, MA 02142-1209, USA journals-info~…}
}

@article{ti2023geometric,
  title={A geometric optimal control approach for imitation and generalization of manipulation skills},
  author={Ti, Boyang and Razmjoo, Amirreza and Gao, Yongsheng and Zhao, Jie and Calinon, Sylvain},
  journal={Robotics and Autonomous Systems},
  volume={164},
  pages = {104413},
  year={2023},
  publisher={Elsevier}
}

@inproceedings{berio2017generating,
            year = {2017},
       booktitle = {Graphics Interface 2017},
          author = {Daniel Berio and Sylvain Calinon and Frederic Fol Leymarie},
           month = {May},
       publisher = {ACM},
           title = {Generating Calligraphic Trajectories with Model Predictive Control}
}

@article{khansari2011learning,
  title={Learning stable nonlinear dynamical systems with gaussian mixture models},
  author={Khansari-Zadeh, S Mohammad and Billard, Aude},
  journal={IEEE Transactions on Robotics},
  volume={27},
  number={5},
  pages={943--957},
  year={2011},
  publisher={IEEE}
}

@article{figueroa2022locally,
  title={Locally active globally stable dynamical systems: Theory, learning, and experiments},
  author={Figueroa, Nadia and Billard, Aude},
  journal={The International Journal of Robotics Research},
  volume={41},
  number={3},
  pages={312--347},
  year={2022},
  publisher={SAGE Publications Sage UK: London, England}
}

@article{lemme2014neural,
  title={Neural learning of vector fields for encoding stable dynamical systems},
  author={Lemme, Andre and Neumann, Klaus and Reinhart, Ren{\'e} Felix and Steil, Jochen J},
  journal={Neurocomputing},
  volume={141},
  pages={3--14},
  year={2014},
  publisher={Elsevier}
}

@inproceedings{nawaz2024learning,
  title={Learning complex motion plans using neural odes with safety and stability guarantees},
  author={Nawaz, Farhad and Li, Tianyu and Matni, Nikolai and Figueroa, Nadia},
  booktitle={2024 IEEE International Conference on Robotics and Automation (ICRA)},
  pages={17216--17222},
  year={2024},
  organization={IEEE}
}

@inproceedings{petrivc2014online,
  title={Online learning of task-specific dynamics for periodic tasks},
  author={Petri{\v{c}}, Tadej and Gams, Andrej and {\v{Z}}lajpah, Leon and Ude, Ale{\v{s}}},
  booktitle={2014 IEEE/RSJ International Conference on Intelligent Robots and Systems},
  year={2014},
  organization={IEEE}
}

@inproceedings{kulak2020fourier,
  title={Fourier movement primitives: an approach for learning rhythmic robot skills from demonstrations.},
  author={Kulak, Thibaut and Silv{\'e}rio, Joao and Calinon, Sylvain},
  booktitle={Robotics: Science and systems},
  year={2020}
}

@article{li2023prodmp,
  title={Prodmp: A unified perspective on dynamic and probabilistic movement primitives},
  author={Li, Ge and Jin, Zeqi and Volpp, Michael and Otto, Fabian and Lioutikov, Rudolf and Neumann, Gerhard},
  journal={IEEE Robotics and Automation Letters},
  volume={8},
  number={4},
  pages={2325--2332},
  year={2023},
  publisher={IEEE}
}

@inproceedings{dixon2005live,
  title={Live tracking of musical performances using on-line time warping},
  author={Dixon, Simon},
  booktitle={Proceedings of the 8th International Conference on Digital Audio Effects},
  volume={92},
  pages={97},
  year={2005},
  organization={Citeseer}
}

@inproceedings{lasota2019bayesian,
  title={Bayesian estimator for partial trajectory alignment},
  author={Lasota, Przemyslaw Andrzej and Shah, Julie A},
  year={2019},
  booktitle={Robotics: Science and Systems},
  year={2019}
}

@article{ravichandar2020recent,
  title={Recent advances in robot learning from demonstration},
  author={Ravichandar, Harish and Polydoros, Athanasios S and Chernova, Sonia and Billard, Aude},
  journal={Annual review of control, robotics, and autonomous systems},
  volume={3},
  pages={297--330},
  year={2020},
  publisher={Annual Reviews}
}

@article{chi2023diffusion,
  title={Diffusion policy: Visuomotor policy learning via action diffusion},
  author={Chi, Cheng and Xu, Zhenjia and Feng, Siyuan and Cousineau, Eric and Du, Yilun and Burchfiel, Benjamin and Tedrake, Russ and Song, Shuran},
  journal={The International Journal of Robotics Research},
  publisher={SAGE Publications Sage UK: London, England},
  year={2023}
}

@InProceedings{pmlr-v164-mandlekar22a,
  title = 	 {What Matters in Learning from Offline Human Demonstrations for Robot Manipulation},
  author =       {Mandlekar, Ajay and Xu, Danfei and Wong, Josiah and Nasiriany, Soroush and Wang, Chen and Kulkarni, Rohun and Fei-Fei, Li and Savarese, Silvio and Zhu, Yuke and Mart\'in-Mart\'in, Roberto},
  booktitle = 	 {Proceedings of the 5th Conference on Robot Learning},
  pages = 	 {1678--1690},
  year = 	 {2022},
  volume = 	 {164},
  series = 	 {Proceedings of Machine Learning Research},
  publisher =    {PMLR}
}

@phdthesis{fourie2024real,
  title={Real-Time Anticipation and Entrainment in Human-Robot Interaction},
  author={Fourie, Christopher},
  year={2024},
  school={Massachusetts Institute of Technology}
}

@inproceedings{sharan2024plan,
  title={Plan Diffuser: Grounding LLM Planners with Diffusion Models for Robotic Manipulation},
  author={Sharan, SP and Zhao, Ruihan and Wang, Zhangyang and Chinchali, Sandeep P and others},
  booktitle={Bridging the Gap between Cognitive Science and Robot Learning in the Real World: Progresses and New Directions},
  year={2024}
}

@INPROCEEDINGS{zhao2023learning, 
    AUTHOR    = {Tony Z. Zhao AND Vikash Kumar AND Sergey Levine AND Chelsea Finn}, 
    TITLE     = {{Learning Fine-Grained Bimanual Manipulation with Low-Cost Hardware}}, 
    BOOKTITLE = {Proceedings of Robotics: Science and Systems}, 
    YEAR      = {2023}, 
    ADDRESS   = {Daegu, Republic of Korea}, 
    MONTH     = {July}, 
    DOI       = {10.15607/RSS.2023.XIX.016} 
}

@inproceedings{urain2020imitationflow,
  title={Imitationflow: Learning deep stable stochastic dynamic systems by normalizing flows},
  author={Urain, Julen and Ginesi, Michele and Tateo, Davide and Peters, Jan},
  booktitle={2020 IEEE/RSJ International Conference on Intelligent Robots and Systems (IROS)},
  pages={5231--5237},
  year={2020},
  organization={IEEE}
}

@inproceedings{calinon2011encoding,
  title={Encoding the time and space constraints of a task in explicit-duration hidden Markov model},
  author={Calinon, Sylvain and Pistillo, Antonio and Caldwell, Darwin G},
  booktitle={2011 IEEE/RSJ International Conference on Intelligent Robots and Systems},
  pages={3413--3418},
  year={2011},
  organization={IEEE}
}

@article{calinon2010learning,
  title={Learning and reproduction of gestures by imitation},
  author={Calinon, Sylvain and D'halluin, Florent and Sauser, Eric L and Caldwell, Darwin G and Billard, Aude G},
  journal={IEEE Robotics \& Automation Magazine},
  volume={17},
  number={2},
  pages={44--54},
  year={2010},
  publisher={IEEE}
}
﻿
﻿
\end{document}